\documentclass[a4paper,fleqn]{cas-sc}

\usepackage[numbers]{natbib}
\usepackage{amsmath,amsfonts}
\usepackage{algorithmic}
\usepackage{algorithm}
\usepackage{amsmath}
\usepackage{amssymb}
\usepackage{array}
\usepackage[caption=false,font=normalsize,labelfont=sf,textfont=sf]{subfig}
\usepackage{textcomp}
\usepackage{stfloats}
\usepackage{url}
\usepackage{verbatim}
\usepackage{graphicx}
\usepackage{pifont}
\newcommand{\cmark}{\ding{51}}  
\newcommand{\xmark}{\ding{55}} 
\usepackage[table]{xcolor}
\usepackage{booktabs,tabularx}
\usepackage{multirow}
\usepackage[utf8]{inputenc}
\usepackage[T1]{fontenc}

\def\tsc#1{\csdef{#1}{\textsc{\lowercase{#1}}\xspace}}
\tsc{WGM}
\tsc{QE}

\begin{document}
\let\WriteBookmarks\relax
\def\floatpagepagefraction{1}
\def\textpagefraction{.001}

\shorttitle{Revisiting  Catastrophic Forgetting in Continual Knowledge Graph Embedding}    

\shortauthors{G. Pons et al.}  

\title [mode = title]{Revisiting  Catastrophic Forgetting in Continual Knowledge Graph Embedding}  

















\author[1]{Gerard Pons}[orcid=0000-0003-2225-3255]
\cormark[1]
\ead{gerard.pons.recasens@upc.edu}
\credit{Conceptualization, Methodology, Software, Writing - original draft}

\author[2]{Carlos Escolano}[orcid=0000-0001-6657-673X]
\ead{carlos.escolano@upc.edu}
\credit{Conceptualization, Methodology, Validation, Visualization, Writing - review \& editing}

\author[1]{Besim Bilalli}[orcid=0000-0002-0575-2389]
\ead{besim.bilalli@upc.edu}
\credit{Supervision, Validation, Writing - review \& editing}

\author[1]{Anna Queralt}[orcid=0000-0003-2782-2955]
\ead{anna.queralt@upc.edu}
\credit{Supervision, Project administration, Writing - review \& editing}

\affiliation[1]{organization={Data-intensive Technologies and Knowledge Systems (DTAK), Department of Service and Information System Engineering, Universitat Politècnica de Catalunya},
            city={Barcelona},
            postcode={08034},
            state={Catalonia},
            country={Spain}}

\affiliation[2]{organization={Intelligent Data Science and Artificial Intelligence Research Center (IDEAI), Department of Computer Science, Universitat Politècnica de Catalunya},
            city={Barcelona},
            postcode={08034},
            state={Catalonia},
            country={Spain}}

\cortext[1]{Corresponding author}



\begin{abstract}
Knowledge Graph Embeddings (KGEs) support a wide range of downstream tasks over Knowledge Graphs (KGs). In practice, KGs evolve as new entities and facts are added, motivating Continual Knowledge Graph Embedding (CKGE) methods that update embeddings over time. Current CKGE approaches address catastrophic forgetting (i.e., the performance degradation on previously learned tasks) primarily by limiting changes to existing embeddings.

However, we show that this view is incomplete. When new entities are introduced, their embeddings can interfere with previously learned ones, causing the model to predict them in place of previously correct answers. This phenomenon, which we call \textit{entity interference}, has been largely overlooked and is not accounted for in current CKGE evaluation protocols. As a result, the assessment of catastrophic forgetting becomes misleading, and CKGE methods performance is systematically overestimated.

To address this issue, we introduce a corrected CKGE evaluation protocol that accounts for entity interference. Through experiments on multiple benchmarks, we show that ignoring this effect can lead to performance overestimation of up to 25\%, particularly in scenarios with significant entity growth. We further analyze how different CKGE methods and KGE models are affected by the different sources of forgetting, and introduce a catastrophic forgetting metric tailored to CKGE.
\end{abstract}


\maketitle

\section{Introduction}
\label{sec:introduction}
Knowledge Graph Embeddings (KGEs) represent the entities and relations of a Knowledge Graph (KG) as vectors in a continuous space. By capturing the underlying structural and semantic properties of the graph, KGEs enable a wide range of downstream tasks over KGs. However, real-world KGs are dynamic and continuously evolve over time through the addition of new entities, relations, and facts~\citep{polleres_hal-04373234} and, as the graph evolves, embeddings quickly become outdated. Consequently, embeddings must be updated to reflect these changes, both by learning embeddings for new entities and relations and by updating existing embeddings to incorporate new knowledge. Efficiently maintaining high-quality embeddings under such evolving conditions remains a key challenge.

\begin{figure}
    \centering
    \includegraphics[width=0.5\textwidth]{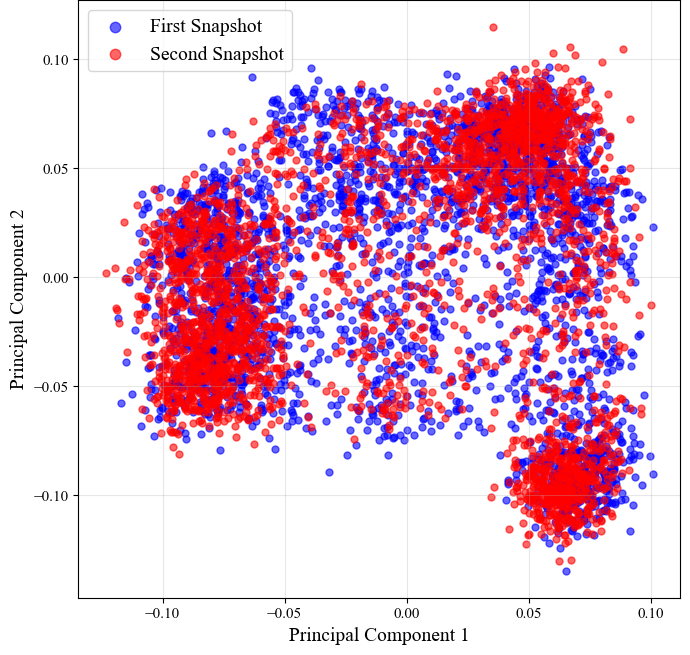}
    \caption{UMAP representation of the embedding space of the ENTITY dataset~\citep{LKGE} for the entities introduced in the first (blue) and second (red) snapshots.}
    \label{fig:dense}
\end{figure}

The Continual Knowledge Graph Embedding (CKGE) field has recently gained traction~\citep{LKGE,incDE,fastKGE,FMR,CMKGE,ETT-CKGE,SAGE,STCKGE,LGD,DebiasedKGE,DyGM}, building upon  continual learning techniques~\citep{continualSurvey} and extending them to the KGE setting. Existing CKGE approaches are designed with the goal of enabling the acquisition of new knowledge while mitigating catastrophic forgetting, which refers to the degradation of performance on previously learned tasks when training with new data. However, current CKGE evaluation protocols do not consider newly introduced entities  when evaluating existing embeddings. As a result, potential prediction errors caused by these entities are not captured, leading to an overestimation of CKGE performance.

We identify this overlooked source of performance degradation as \textit{entity interference}. Entity interference arises from a fundamental characteristic of CKGE: the need to incorporate new entity embeddings resulting from the evolution of the underlying KG. For instance, in Wikidata~\citep{wikidata}, from 2018 to 2021 the number of entities increased from 40M to 90M. This requires the creation of new embeddings, often resulting in denser embedding spaces. Figure~\ref{fig:dense} illustrates this phenomenon for a widely used CKGE benchmark dataset~\citep{LKGE}, where the embeddings for previously learned entities (blue) and for the newly introduced ones (red) are in the same regions of the embedding space. This is an expected behavior, since an embedding model that is able to capture the underlying semantics of the graph, should place similar entities (e.g., those sharing classes) close to each other in the embedding space. However, the increased density of the embedding space poses a crucial challenge to CKGE methods, as the embeddings of newly introduced entities may interfere with existing ones, causing predictions that were previously correct to become incorrect when these new entities are mistakenly predicted in their place.

Despite its potential impact, entity interference has been largely overlooked in the design of catastrophic forgetting mitigation strategies, as existing approaches mainly attempt to reduce forgetting by limiting undesired changes in previously learned embeddings. We refer to this source of forgetting as \textit{representation drift}.

Ignoring entity interference in current evaluation protocols leads to a significant overestimation of CKGE performance. When new entities are excluded from the evaluation candidate set, reported results can be up to 25\% higher than those obtained when they are properly considered. Moreover, this omission creates the false impression of effective catastrophic forgetting mitigation. In particular, CKGE methods that completely prevent changes in existing embeddings may appear to achieve perfect catastrophic forgetting mitigation, even though performance loss still occurs due to interference from newly introduced entities.

In this paper, we demonstrate that entity interference is a major source of catastrophic forgetting in CKGE and should be accounted for during evaluation, as its exclusion can substantially overestimate model performance. We then analyze in depth the different sources of catastrophic forgetting and their impact on both CKGE methods and KGE models. Finally, we introduce a unified metric for quantifying catastrophic forgetting, addressing the lack of consistent evaluation measures in prior work.

Thus, the main contributions of this work are as follows:
\begin{itemize}
    \item We identify entity interference as an overlooked phenomenon in CKGE, which has been excluded from the evaluation protocols and has an important impact on catastrophic forgetting.
    \item We introduce a corrected CKGE evaluation protocol, and conduct extensive experiments on a large set of state-of-the-art datasets identifying scenarios in which the incomplete evaluation leads to significant overestimation of model’s performance.
    \item We propose a unified metric to quantify catastrophic forgetting in CKGE, addressing the lack of consistent metrics in prior work.
    \item We analyze how different CKGE methods mitigate catastrophic forgetting and evaluate how they are affected by it, examining the contribution of the different sources of forgetting.
    \item We evaluate the effects that the different sources of catastrophic forgetting have on KGE models, examining the trade-off between forgetting and knowledge acquisition.
\end{itemize}

The remainder of this paper is organized as follows. Section~\ref{sec:continual} introduces the preliminaries of CKGE and positions them within the continual learning taxonomy. Section~\ref{sec:catastrophic} characterizes catastrophic forgetting in CKGE, decomposing it into two distinct sources. Then, Section~\ref{sec:limitations} outlines and corrects the limitations in CKGE evaluation and proposes an appropriate catastrophic forgetting metric for CKGE. Next, Section~\ref{sec:forgetting_mitigation} categorizes existing catastrophic forgetting mitigation techniques. Section~\ref{sec:experiments} presents an experimental evaluation of different CKGE methods and KGE models, analyzing both the impact of the corrected evaluation protocol and the contributions of the different sources of forgetting. Finally, we provide conclusions and directions for future work in Section~\ref{sec:conclusions}.

\section{Continual Learning of KGE}
In standard machine learning, models are typically trained under the assumption of a static dataset. However, many real-world applications operate in dynamic environments where data arrive sequentially over time. Continual learning addresses this setting by studying how models can incrementally acquire and refine knowledge while limiting the degradation of previously learned information, known as catastrophic forgetting. 

This dynamic setting is also found in KGs, as they evolve over time through the addition of new entities, relations, and facts. Consequently, their KGEs, which are learned via a machine learning model, must also undergo continual updates to preserve and refine existing knowledge while integrating new information. In the following subsections, we introduce the preliminaries of CKGE and position the problem within the established continual learning scenarios.
\label{sec:continual}
\subsection{Preliminaries}
\label{preliminaries}

\paragraph*{Growing Knowledge Graphs} A KG is a collection of triples $\mathcal{S} \subseteq \mathcal{E} \times \mathcal{R} \times \mathcal{E}$, where $\mathcal{E}$ and $\mathcal{R}$ denote the sets of entities and relations, respectively. Each triple $(h,r,t)$ is composed of head and tail entities $h,t \in \mathcal{E}$ and a relation $r \in \mathcal{R}$. The evolution of a KG can be represented as a sequence of snapshots $\mathcal{KG} = \{ \mathcal{S}_0, \mathcal{S}_1, \ldots, \mathcal{S}_n\}$, where each snapshot expands on the previous one, i.e., $\mathcal{S}_{i-1} \subseteq \mathcal{S}_i$, where $\mathcal{E}_{i-1} \subseteq \mathcal{E}_i$, and $\mathcal{R}_{i-1} \subseteq \mathcal{R}_i$. The newly introduced triples, entities, and relations at snapshot $i$ are denoted as $\Delta \mathcal{S}_i = \mathcal{S}_i \setminus \mathcal{S}_{i-1}$, $\Delta \mathcal{E}_i = \mathcal{E}_i \setminus \mathcal{E}_{i-1}$, and $\Delta \mathcal{R}_i = \mathcal{R}_i \setminus \mathcal{R}_{i-1}$, respectively. 

\paragraph*{Knowledge Graph Embedding} KGEs are low-dimensional representations of the entities and relations of a KG. They are represented as d-dimensional vectors, being $\mathbf{e} \in \mathbb{R}^d$ and $\mathbf{r} \in \mathbb{R}^d$ the embeddings of an entity $e$ and a relation $r$, respectively, and they are organized into the corresponding $\textbf{E}\in \mathbb{R}^{|\mathcal{E}|\times d}$ and $\textbf{R}\in \mathbb{R}^{|\mathcal{R}|\times d}$ matrices. Embeddings are learned by training a KGE model $\mathcal{M}$, which updates the values of the embeddings by minimizing the model's loss function $\mathcal{L}_{\mathcal{M}}$. To do so, the triples in the set $\mathcal{S}$ are divided into a training set $\mathcal{T}$ and a test set $\mathcal{G}$.  The model's loss is designed so that, once optimized, the model can discriminate between more and less plausible triples. For instance, in the simplest translation-based KGE model TransE~\citep{TransE}, the embeddings are viewed as translations from the head entity to the tail entity via the relation. Therefore, in this case, for a given triple $(h,r,t)$, the model would minimize $\mathcal{L}_{\mathcal{M}}(h, r, t)=\mathbf{h}+\mathbf{r}-\mathbf{t}$, assigning a higher plausibility score $p_{\mathcal{M}}(h, r, t)$ to triples that more closely satisfy this translation.

\paragraph*{Continual Knowledge Graph Embedding}
In the CKGE setting, embeddings $\textbf{E}_0\in \mathbb{R}^{|\mathcal{E}_0|\times d}$ and $\textbf{R}_0\in \mathbb{R}^{|\mathcal{R}_0|\times d}$ are first produced for the entities $\mathcal{E}_0$ and relations $\mathcal{R}_0$ of the first snapshot $\mathcal{S}_0$, by training a KGE model $\mathcal{M}$ with $\mathcal{T}_0$. When new triples arrive at $\Delta\mathcal{S}_i$, embeddings must be learned for the newly introduced entities $\Delta \mathcal{E}_i$ and relations $\Delta \mathcal{R}_i$, while the embeddings of existing entities  $\mathcal{E}_{i-1}$ and relations $\mathcal{R}_{i-1}$ need to be updated, producing $\mathbf{E}_{i+1}$ and $\mathbf{R}_{i+1}$. This can be achieved by using common continual learning methods, such as finetuning, by modifying the model's architecture, or by expanding $\mathcal{L}_{\mathcal{M}}$ with regularization or reconstruction terms, as explained in Section~\ref{sec:catastrophic}. For that, the incremental training and test set, $\Delta\mathcal{T}_i$ and $\Delta\mathcal{G}_i$ respectively, are used.

\subsection{Understanding CKGE Beyond Standard Continual Learning Scenarios}
Various studies have proposed taxonomies to categorize the different scenarios addressed in continual learning~\citep{continualSurvey,typesofC}. Given a task (e.g., classify between A and B) and some input data, these scenarios mainly include: \textit{instance-incremental learning}, where new batches of data arrive for the same task; \textit{domain-incremental learning}, where the data distribution changes over time while the task remains unchanged; \textit{class-incremental learning}, where the model incrementally learns to distinguish between new classes (e.g., classify between A, B, C and D); and \textit{task-incremental learning}, where the model incrementally learns distinct tasks and is informed at inference time which task to solve (e.g, task 1: classify between A and B; task 2: classify between C and D).

In this section, we analyze how CKGE relate to these scenarios and show that they do not fit clearly into a single category. Instead, CKGE simultaneously exhibit characteristics of multiple continual learning settings.

KGEs are typically trained to solve the Link Prediction task~\citep{KGAPP}, where the model assigns a plausibility score $p_{\mathcal{M}}(h, r, t)$ to a triple $(h,r,t)$ and is evaluated on whether it ranks this triple above its corrupted versions. Corrupted triples are those triples that are obtained by replacing either the head (i.e., $h'$) or the tail (i.e., $t'$) entity from the initial triple, without resulting in another triple present in the KG.  When new triples $\Delta\mathcal{S}_i$ arrive, CKGE could be assimilated to \textit{instance-incremental learning}, as these new triples refine the existing embeddings to improve predictive performance. Additionally, depending on the parts of the graph these new triples cover, it may also be interpreted as \textit{domain-incremental learning}. 

More importantly, when new triples introduce new entities $\Delta\mathcal{E}_i$, the model must learn their embeddings and consider them as potential candidates during prediction. This introduces a \textit{class-incremental learning} dimension, particularly in scenarios where the number of entities grows substantially over time. 

However, existing CKGE evaluation protocols do not reflect this class-incremental aspect. Instead, they evaluate past test triples considering only the entities available at the time those triples were observed. As a result, each snapshot is treated as an independent task, aligning the evaluation with a \textit{task-incremental} setting.

This mismatch between the CKGE problem, which involves a growing set of candidate entities, and its evaluation protocol has important consequences. In particular, it prevents capturing the interference introduced by newly added entities. As we show in Section~\ref{sec:understanding}, this mismatch plays a central role in the overestimation of model performance.

\section{Sources of Catastrophic Forgetting in CKGE}
\label{sec:catastrophic}

In this section, we define catastrophic forgetting in CKGE and show that it arises from two distinct sources: representation drift and entity interference.

\subsection{Defining Catastrophic Forgetting in CKGE}
\label{sec:understanding}
\begin{figure}
    \centering
    \includegraphics[width=0.9\linewidth]{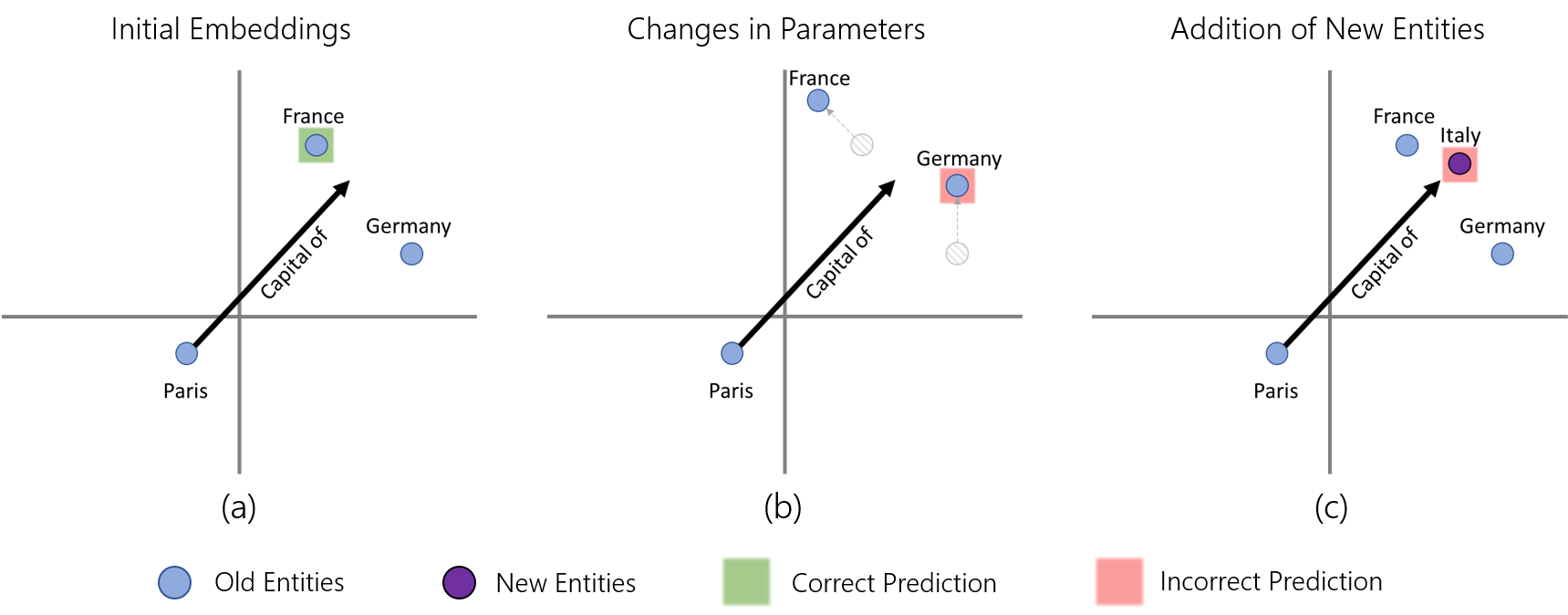}
    \caption{In (a) an initial correct TransE Link Prediction example. In (b), after a continual learning process the embeddings have shifted and the result of link prediction is incorrect. In (c), a new entity interferes with the previously learned knowledge and causes an incorrect prediction.}
    \label{fig:catastrophic}
\end{figure}

In the context of KGE, the knowledge learned by a model $\mathcal{M}$ is typically evaluated on the Link Prediction task. Formally, we say that the knowledge of a triple $(h,r,t)$ is correctly captured at snapshot $i$, if for all corrupted triples $(h',r,t)$ and $(h,r,t')$: 

\begin{equation}\label{eq:score}
\begin{aligned}
p_{\mathcal{M}}(\mathbf{h}',\mathbf{r},\mathbf{t}) &<
p_{\mathcal{M}}(\mathbf{h},\mathbf{r},\mathbf{t})\;\;
\forall\, h'\in\mathcal{E}_i\setminus\{h\}:\; (h',r,t)\notin\mathcal{S}_i \\[2pt]
& \qquad \qquad \qquad \text{and} \qquad \qquad \qquad\\[2pt]
\hspace{0.5em}p_{\mathcal{M}}(\mathbf{h},\mathbf{r},\mathbf{t}') &<
p_{\mathcal{M}}(\mathbf{h},\mathbf{r},\mathbf{t})\;\;
\forall\, t' \in \mathcal{E}_i\setminus\{t\}:\; (h,r,t') \notin \mathcal{S}_i
\end{aligned}
\end{equation}

This can be observed in Figure~\ref{fig:catastrophic} (a) for tail prediction in the KGE model TransE. In this simple model, given an incomplete triple $(h,r,?)$, the most plausible tail entity is the one whose embedding is closest to the vector obtained by adding the head and relation embeddings. For example, adding the embedding of \textit{Paris} to that of the relation \textit{capital of} yields a point in the embedding space whose closest embedding is the one of the entity \textit{France}.

After training on the new data $\Delta \mathcal{T}_{i+1}$ at snapshot $i+1$, if Equation~\ref{eq:score} no longer holds for an entity $e \in \mathcal{E}_i$ for which it previously held, we consider this an occurrence of catastrophic forgetting. Existing work typically attributes catastrophic forgetting to undesired drift in the embeddings of previously learned entities $\mathbf{E}_i$ and relations $\mathbf{R}_i$. This view is inherited from standard continual learning settings which, as discussed in Section~\ref{sec:continual}, do not directly map to the CKGE scenario. In the CKGE context, however, we show that catastrophic forgetting is not only caused by drift in existing embeddings but can also arise from the interference introduced by newly added entities. Consequently, catastrophic forgetting in CKGE originates from two distinct sources.

\subsection{Representation Drift}

During the continual learning process, embeddings of previously learned entities $\mathcal{E}_i$ and relations $\mathcal{R}_i$ may change. Such changes can be desirable; for instance, a new triple $(\text{France}, \text{located in}, \text{Europe})$ may enrich the embeddings of France and Europe. However, these updates can also erase previously acquired knowledge. We refer to this phenomenon as catastrophic forgetting caused by representation drift. That is, if we only consider the updated embeddings $\mathbf{E}_{i+1}[\mathcal{E}_i] = \{\, \mathbf{E}_{i+1}[e, :] \mid e \in \mathcal{E}_i \,\}$ and $\mathbf{R}_{i+1}[\mathcal{R}_i] = \{\, \mathbf{R}_{i+1}[r, :] \mid r \in \mathcal{R}_i \,\}$, some of the knowledge contained in the previous triples $\mathcal{S}_i$ has been lost. Formally, we say that representation drift forgetting occurs for a triple $(h,r,t)\in\mathcal{S}_i$ previously satisfying the conditions in Equation~\ref{eq:score} when: 


\begin{equation}
\begin{split}
& \exists\,h^\prime \in \mathcal{E}_{i}, \; h^\prime \neq h, \; (h',r,t) \notin \mathcal{S}_{i+1} :  p_{\mathcal{M}}(\mathbf{h'}, \mathbf{r}, \mathbf{t}) >
p_{\mathcal{M}}(\mathbf{h}, \mathbf{r}, \mathbf{t}) \\[2pt]
& \hspace{4cm} \text{or} \\[2pt]
& \exists\,t^\prime \in \mathcal{E}_{i}, \; t^\prime \neq t, \; (h,r,t') \notin \mathcal{S}_{i+1} :  p_{\mathcal{M}}(\mathbf{h}, \mathbf{r}, \mathbf{t'}) >
p_{\mathcal{M}}(\mathbf{h}, \mathbf{r}, \mathbf{t})
\end{split}
\end{equation}

where all the embeddings are taken from the updated embedding matrices $\mathbf{E}_{i+1}$ and $\mathbf{R}_{i+1}$. This is exemplified in Figure~\ref{fig:catastrophic} (b), where after the continual learning stage, the embeddings of the entities \textit{France} and \textit{Germany} have shifted, resulting in the entity with the highest score for the incomplete triple (\textit{Paris}, \textit{capital of}, ?) being \textit{Germany}.

\subsection{Entity Interference}

Changes in existing embeddings are not the only cause for the loss of knowledge. The embeddings learned for newly introduced entities $\Delta \mathcal{E}_{i+1}$ may be positioned close to the embeddings of existing entities $\mathcal{E}_i$, which can lead the embedding model to produce incorrect predictions. This is exemplified in Figure~\ref{fig:catastrophic} (c), where although the embeddings for existing entities remain unchanged, the embedding learned for the new entity  \textit{Italy} interferes with the prediction  (\textit{Paris}, \textit{capital of}, ?). Formally, we say that entity interference has occurred for given a triple $(h,r,t)\in\mathcal{S}_i$ previously satisfying the conditions in Equation~\ref{eq:score} when: 


\begin{equation}
\begin{split}
& \exists\,h^\prime \in \Delta\mathcal{E}_{i+1}, \; h^\prime \neq h, \; (h',r,t) \notin \mathcal{S}_{i+1} :  p_{\mathcal{M}}(\mathbf{h'}, \mathbf{r}, \mathbf{t}) > p_{\mathcal{M}}(\mathbf{h}, \mathbf{r}, \mathbf{t}) \\[2pt]
& \hspace{4cm} \text{or} \\[2pt]
& \exists\,t^\prime \in \Delta\mathcal{E}_{i+1}, \; t^\prime \neq t, \; (h,r,t') \notin \mathcal{S}_{i+1} : p_{\mathcal{M}}(\mathbf{h}, \mathbf{r}, \mathbf{t'}) > p_{\mathcal{M}}(\mathbf{h}, \mathbf{r}, \mathbf{t})
\end{split}
\end{equation}

where all the embeddings are taken from the updated embedding matrices $\mathbf{E}_{i+1}$ and $\mathbf{R}_{i+1}$. 

While representation drift has been widely studied in CKGE, the impact of entity interference has received little to no attention. 
\section{Limitations of Current CKGE Evaluation}
\label{sec:limitations}
KGEs are evaluated with the Link Prediction task. In this setting, the metrics used (e.g., Hits@k and Mean Reciprocal Rank) are derived from the position (i.e., rank) of the correct entity when assessing the scores given to an incomplete triple (h, r, ?):



\begin{equation}
\mathrm{rank}(h,r,t) = 1 + \Bigl|
\bigl\{ t' \in \mathcal{E} \,\big|\,
p_{\mathcal{M}}(\mathbf{h},\mathbf{r},\mathbf{t}') >
p_{\mathcal{M}}(\mathbf{h},\mathbf{r},\mathbf{t}),
\ (h,r,t') \notin \mathcal{S}
\bigr\}
\Bigr|
\label{eq:rank}
\end{equation}

Thus, a key question is which entity set should be considered when evaluating past test triples $\Delta\mathcal{G}_j$ after training on a later snapshot $i$. Existing CKGE methods~\citep{LKGE,incDE,fastKGE,FMR,ETT-CKGE,SAGE,DebiasedKGE} typically restrict the evaluation to the entities available at the time the test triples were observed, i.e., they compute Equation~\ref{eq:rank} using $t' \in \mathcal{E}_j$.

However, this choice ignores the entities introduced in subsequent snapshots. As a result, the evaluation does not capture cases in which newly added entities would be ranked above the correct one. For instance, in Figure~\ref{fig:catastrophic} (c), the prediction for the incomplete triple (\textit{Paris}, \textit{capital of}, ?) would still be attributed to \textit{France} under this protocol, even though \textit{Italy} receives a higher plausibility score in the current embedding space.

This evaluation setting resembles a \textit{task-incremental} scenario, where each snapshot is treated as an isolated task. However, this assumption does not hold in evolving knowledge graphs, where predictions are made over an expanding set of entities. Consequently, the reported results are systematically overestimated, i.e., $\theta_{j,i} \geq \theta_{k,i}\;\; \forall j<k$, where $\theta_{j,i}$ denotes the evaluation metric for the test set $\Delta\mathcal{G}_i$ considering entities in $\mathcal{E}_j$.

\paragraph*{Corrected Evaluation Protocol} When evaluating CKGE by computing the rank (Equation~\ref{eq:rank}), the set of candidate entities should correspond to the current entity set $\mathcal{E}_i$ for all test snapshots, regardless of the snapshot from which the test triple originates. This ensures that evaluation reflects current embedding space. In Section~\ref{sec:exp_correct}, we evaluate existing CKGE techniques under this corrected protocol, showing the impact of the overestimation.

\subsection{Quantifying Catastrophic Forgetting in CKGE}
\label{sec:quantify}

Metrics used to assess catastrophic forgetting are typically derived from any of the previously introduced performance measures $\theta$ evaluated at different stages of training, as they quantify the loss in predictive performance. However, given that computation of $\theta$ does not account for entity interference, these metrics fail to reflect the actual performance degradation. As a result, highly regularized methods, or even those that freeze embeddings, tend to obtain optimal or near-optimal results. As will be shown in Section~\ref{sec:exp_correct}, this is particularly misleading in continual learning settings where the number of entities increases substantially across snapshots. 

Additionally, in the context of CKGE, existing studies adopt heterogeneous approaches to assess catastrophic forgetting (see last column in Table~\ref{tab:ckge}). Some works do not evaluate forgetting explicitly and rely solely on standard KGE Link Prediction metrics (e.g., MRR and Hits@k), making it difficult to assess whether the observed improvements originate from reduced  forgetting or from enhanced acquisition of new knowledge. Other studies assess catastrophic forgetting qualitatively, typically through visualizations that depict trends in knowledge retention. Although these analyses allow to assess overall model behavior, they do not enable straightforward comparisons across methods. Finally, several works use diverse quantitative metrics to measure catastrophic forgetting. 

One of the first CKGE methods~\citep{LKGE} adopts the Backward Transfer (BWT) metric from~\citep{GEM}:
\begin{equation}
    BWT = \frac{1}{N-1}\sum_{i=0}^{N-1}\theta_{N,i}-\theta_{i,i}
    \label{eq:bwt}
\end{equation}
where $\theta_{i,j}$ is the evaluation of a Link Prediction metric $\theta$ (e.g., MRR) for the test set $\Delta\mathcal{G}_i$ after training with snapshot $j$. This metric has been adapted in different CKGE works. For instance, ~\citep{rethinking} argues that the performance on past snapshots may increase overtime, and adapt the summation in Equation~\ref{eq:bwt}'s to $\max_{t \in \{0 \ldots N-1\}} (\theta_{t,i} - \theta_{N,i})$. In~\citep{SAGE}, instead of measuring absolute degradation, relative performance changes are quantified by using the term $\frac{\theta_{N,i}}{\theta_{N,i}+\theta_{,i}}$, while~\citep{ERPP} weights test sets based on their number of triples. It is important to note that, as discussed above, existing methods do not include new entities when computing $\theta_{i,j} \text{ where } j>i$. Thus, catastrophic forgetting is measured only with respect to representation drift over previously learned entities, and does not measure the interference caused by the new ones. 

Additionally, existing metrics assign equal importance to all test snapshots, regardless of how many continual learning steps each snapshot has undergone. This introduces a bias, as earlier snapshots are exposed to more rounds of updates and thus more opportunities for both drift and interference. For example, in a setting with 5 snapshots, the forgetting measured on the first test set, $\theta_{5,0} - \theta_{0,0}$, is weighted the same as the forgetting on the fourth test set, $\theta_{5,4} - \theta_{4,4}$, even though the former has been exposed to four updates while the latter has only been exposed to one. By treating all snapshots equally, the cumulative effect of multiple learning steps on earlier snapshots is overlooked. 

To address these limitations, we propose the following adapted metric:

\begin{equation}
    CF \;=\; \sum_{i=0}^{N-1}
    \left(
      \frac{\theta_{N,i}-\theta_{i,i}}{(N-i)\cdot\theta_{i,i}}
      \cdot
      \frac{|\Delta\mathcal{G}_i|}{\,|\mathcal{G}_N|-|\Delta\mathcal{G}_N|\,}
    \right)
    \label{eq:cf}
\end{equation}

where the term $N-i$ adjusts de contribution of each test set according to the number of continual learning cycles it has undergone, the normalization by $\theta_{i,i}$ captures relative performance degradation, and the weighting by $|\Delta\mathcal{G}_i|$ accounts for differences in test set size, as in~\citep{ERPP}.

In Section~\ref{sec:exp_decompose}, we reevaluate existing CKGE methods using the proposed catastrophic forgetting metric, showing that methods previously reported to exhibit little or no forgetting are, in fact, significantly affected by it.
\section{Mitigating Catastrophic Forgetting in CKGE}
\label{sec:forgetting_mitigation}

\begin{table*}[!t]
\caption{Comparison of CKGE Methods Depending On the Type of Catastrophic Forgetting Mitigation Techniques Implemented and how They Evaluate It.}
\centering
\begin{tabular}{c|c|c|c|c|c|c|c|c|c}
     \textbf{CKGE method} & \textbf{KGE} &\textbf{Replay} &\textbf{Ordering} &\textbf{Archit.} &\textbf{Reg.} &\textbf{Reconst.} &\textbf{Align.} &\textbf{Masking} &\textbf{CF Evaluation} \\
\toprule
LKGE~\citep{LKGE} & TransE & \xmark & \xmark  & \xmark & \cmark  & \cmark & \xmark  & \xmark & Quantitative \\
\midrule
incDE~\citep{incDE} & TransE & \xmark & \cmark  & \cmark & \cmark  & \xmark & \xmark  & \xmark & Qualitative \\
\midrule
FastKGE~\citep{fastKGE} & TransE & \xmark & \xmark  & \cmark & \xmark  & \xmark & \xmark  & \xmark & \xmark \\
\midrule
FMR~\citep{FMR} & TransE & \xmark & \xmark  & \xmark & \cmark  & \cmark & \xmark  & \xmark & Quantitative \\
\midrule
CMKGE~\citep{CMKGE} & TransE & \xmark & \xmark  & \xmark & \cmark  & \cmark & \xmark  & \cmark & Qualitative \\
\midrule
ETT-CKGE~\citep{ETT-CKGE} & TransE & \xmark & \xmark  & \cmark & \cmark  & \xmark & \xmark  & \xmark & Qualitative \\
\midrule
SAGE~\citep{SAGE} & TransE & \cmark & \xmark  & \cmark & \cmark  & \xmark & \xmark  & \xmark & Quantitiative \\
\midrule
STCKGE~\citep{STCKGE} & BoxE & \xmark & \xmark  & \cmark & \cmark  & \xmark & \xmark  & \xmark & Quantitiative \\
\midrule
LGD~\citep{LGD} & TransE & \xmark & \cmark  & \xmark & \cmark  & \xmark & \xmark  & \xmark & \xmark \\
\midrule
CMJA~\citep{CMJA} & TransE & \xmark & \cmark  & \xmark & \cmark  & \cmark & \xmark  & \xmark & Qualitative \\
\midrule
DebiasedKGE~\citep{DebiasedKGE} & TransE & \xmark & \xmark  & \xmark & \cmark  & \cmark & \cmark  & \xmark & Qualitative \\
\midrule
DyGM~\citep{DyGM} & TransE & \cmark & \cmark  & \cmark & \cmark  & \xmark & \xmark  & \xmark & Quantitative

\label{tab:ckge}
\end{tabular}
\end{table*}

\begin{figure}
    \centering
    \includegraphics[width=0.6\linewidth]{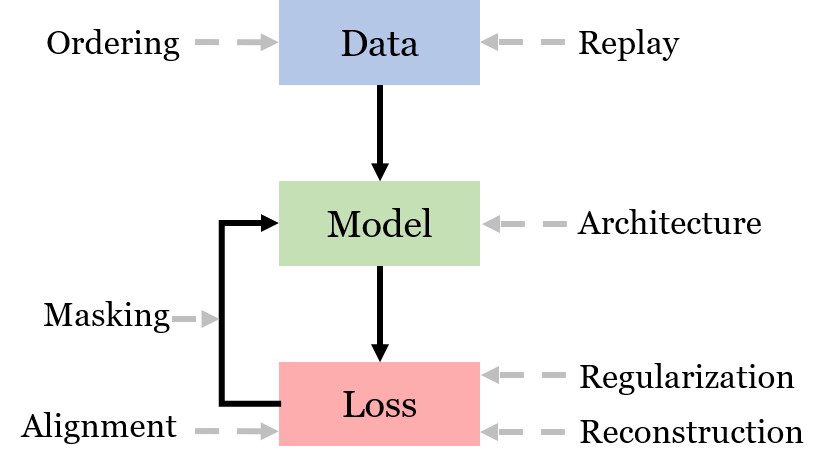}
    \caption{Illustration of where different catastrophic forgetting mitigation techniques are implemented in the learning cycle of CKGE.}
    \label{fig:mitigating}
\end{figure}

In this section, we adapt and extend the standard categorization of catastrophic forgetting mitigation techniques~\citep{continualSurvey} to the CKGE setting, and analyze which sources of forgetting these methods address. To this end, we introduce a higher-level grouping of strategies based on when they intervene during the training cycle. As shown in Figure~\ref{fig:mitigating},  these strategies can operate at different stages: at the data level (e.g., replay and ordering), at the model level (architectural approaches), at the loss level (e.g., regularization, reconstruction, and alignment), or at the weight update level (e.g., masking). Table~\ref{tab:ckge} summarizes the CKGE methods considered in this work, together with their corresponding mitigation strategies, and their approach to the evaluation of catastrophic forgetting. The techniques included are those present in CKGE methods, as well as standard continual learning techniques that have been adapted to the CKGE setting~\citep{LKGE}. Later in Section~\ref{sec:exp_decompose}, we study how these techniques affect the different sources of forgetting.

\subsection{Data Level}
At the data level, the focus is on which data (i.e., the KG triples used to train the KGEs) is used and how it is fed to the model. Two main strategies fall into this category:
\paragraph{Replay} This strategy consists in allocating a memory buffer to store triples from past snapshots (i.e., episodic memory), which are then combined with new triples during training~\citep{GEM,EMR,DyGM}. By replaying past triples, the objective is to preserve previously learned representations by reinforcing past knowledge during gradient updates.
\paragraph{Ordering} This strategy, proposed in~\citep{incDE}, orders newly observed triples according to their proximity to the existing KG triples using breadth-first search, and feeds them to the model according to this order. For instance, in snapshot $i$, the first triples will be those composed by entities and relations from $\mathcal{E}_{i-1}$ and $\mathcal{R}_{i-1}$, followed by those triples which have just one new entity or relation. Inside each partition, triples are then further ordered based on graph metrics, namely their node and betweenness centrality. With this learning order, the aim is to reduce the disruption to existing embeddings.  

\subsection{Model level}
At the model level, the strategies are related to how the \textit{Architecture} of the KGE models is adapted over time. One approach, proposed in~\citep{SAGE}, expands the embedding dimension $d$ when training with new data in order to better accommodate new knowledge. Another strategy consists of freezing  old entity embeddings (i.e., preventing them from being updated)~\citep{fastKGE}, which completely avoids the forgetting originated by representation drift but does not allow new knowledge to be integrated within them. In contrast, ~\citep{incDE, DyGM} adopt a multi-stage procedure where embeddings are frozen during an initial phase to reduce abrupt changes while new embeddings are being learned, and are then unfrozen to allow joint refinement.

\subsection{Loss level}
At the loss level, catastrophic forgetting is mitigated by adding new terms to the loss function of the embedding model $\mathcal{L}_{\mathcal{M}}$. These additional terms can serve different purposes:
\paragraph{Regularization} This technique consists of penalizing changes in past embeddings:

\begin{equation}
    \mathcal{L}_{reg}=\sum_{e\in\mathcal{E}_{i-1}}w(e)\,\Delta_{\Psi}(\mathbf{e}_i,\mathbf{e}_{i-1}) + \sum_{r\in\mathcal{R}_{i-1}}w(r)\,\Delta_{\Psi}(\mathbf{r}_i,\mathbf{r}_{i-1})
\end{equation}
where $w(\cdot)$ is the weight given to each embedding and $\Delta_{\Psi}$ denotes the penalty given to the changes in the embeddings according to $\Psi$ (e.g., the norm of the difference). Generic regularized continual learning approaches use as $w(\cdot)$ the Fisher information matrix~\citep{EWC} or  the past improvements over the loss by changes in that parameter~\citep{SI}. In the CKGE setting some works~\citep{LKGE, FMR, SAGE, CMJA} use the number of new and old triples containing each entity to compute the weights. Other CKGE approaches leverage KG metrics such as centrality of the embeddings ~\citep{incDE}, or use trainable parameters~\citep{ETT-CKGE}.
\paragraph{Reconstruction} 
To mitigate knowledge instability when updating the embeddings, some methods reconstruct the embeddings based on the representations of their neighboring entities and the relations linking them:

\begin{equation}
\mathcal{L}_{\text{rec}} =
\sum_{e \in \mathcal{E}_{i}}  \lVert \mathbf{e}_i - \mathbf{\tilde{e}}_i \rVert_2^2
\;+\;
\sum_{r \in \mathcal{R}_{i}} \lVert \mathbf{r}_i - \mathbf{\tilde{r}}_i \rVert_2^2
\end{equation}

To produce the reconstructed embeddings  $\mathbf{\tilde{e}}$ and $\mathbf{\tilde{r}}$, some works apply Graph Convolutional Networks with node aggregation~\citep{FMR,DebiasedKGE,CMKGE}, while others obtain them by using the underlying structural assumptions of the KGE model (e.g., in TransE, for a triple (h, r, t), the reconstructed embedding $\mathbf{\tilde{e}}_t$ can be obtained as $\mathbf{e}_h+\mathbf{e}_r$)~\citep{LKGE, CMJA}.

\paragraph{Alignment} In~\citep{DebiasedKGE}, it is argued that under representation drifts, past knowledge is not necessarily lost but can become misaligned due to semantic drift in the embedding space, particularly in relation embeddings. To address this, additional alignment terms are introduced into the loss function to constrain the similarity between new and old embeddings. In particular, a cosine-based alignment loss is used to minimize directional changes.

\begin{equation}
\mathcal{L}_{\text{align}} =
\sum_{e \in \mathcal{E}_{i}}  (1 -cos(\mathbf{e}_i,\mathbf{e}_{i-1}))
\;+\;
\sum_{r \in \mathcal{R}_{i}} (1 -cos(\mathbf{r}_i,\mathbf{r}_{i-1}))
\end{equation}

\subsection{Weight Update Level}
Finally, at the weight update level, \textit{Masking} has been proposed as a way to reduce catastrophic forgetting~\citep{CMKGE}. Concretely, a mask is defined to decide which parts of old embeddings should not change.

As has been seen, regardless of the level in which they operate, existing catastrophic forgetting mitigation strategies in CKGE are primarily designed to address representation drift, either by restricting parameter updates, by reconstructing prior embeddings, or by enforcing alignment across snapshots. As a result, entity interference arising from the introduction of new entities remains unaddressed. In Section~\ref{sec:exp_decompose}, we analyze how these techniques affect the different sources of catastrophic forgetting.
\section{Experiments}
\begin{table*}[ht!]
\caption{Summary of the Dataset Statistics Indicating Cumulative Number of Entities, Cumulative Number of Relations and New Triples at Each Snapshot.}
\label{tab:datasets}
\centering
\small
\renewcommand{\arraystretch}{1.25}
\resizebox{\textwidth}{!}{
\begin{tabular}{l|ccc|ccc|ccc|ccc|ccc}
\hline
 &
\multicolumn{3}{c|}{\textbf{Snapshot 0}} &
\multicolumn{3}{c|}{\textbf{Snapshot 1}} &
\multicolumn{3}{c|}{\textbf{Snapshot 2}} &
\multicolumn{3}{c|}{\textbf{Snapshot 3}} &
\multicolumn{3}{c}{\textbf{Snapshot 4}} \\
\textbf{Dataset} &
$|E_0|$ & $|R_0|$ & $|\Delta S_0|$ &
$|E_1|$ & $|R_1|$ & $|\Delta S_1|$ &
$|E_2|$ & $|R_2|$ & $|\Delta S_2|$ &
$|E_3|$ & $|R_3|$ & $|\Delta S_3|$ &
$|E_4|$ & $|R_4|$ & $|\Delta S_4|$ \\
\hline
ENTITY      & 2{,}909 & 233 & 46{,}388 & 5{,}817 & 236 & 72{,}111 & 8{,}275 & 236 & 73{,}785 & 11{,}633 & 237 & 70{,}506 & 14{,}541 & 237 & 47{,}326 \\
RELATION    & 11{,}560 & 48 & 98{,}819 & 13{,}343 & 96 & 93{,}535 & 13{,}754 & 143 & 66{,}136 & 14{,}387 & 190 & 30{,}032 & 14{,}541 & 237 & 21{,}594 \\
FACT        & 10{,}513 & 237 & 62{,}024 & 12{,}779 & 237 & 62{,}023 & 13{,}586 & 237 & 62{,}023 & 13{,}894 & 237 & 62{,}023 & 14{,}541 & 237 & 62{,}023 \\
HYBRID      & 8{,}628 & 86 & 57{,}561 & 10{,}040 & 102 & 20{,}873 & 12{,}779 & 151 & 88{,}017 & 14{,}393 & 209 & 103{,}339 & 14{,}541 & 237 & 40{,}326 \\
GraphEqual  & 2{,}908 & 226 & 57{,}636 & 5{,}816 & 235 & 62{,}023 & 8{,}724 & 237 & 62{,}023 & 11{,}632 & 237 & 62{,}023 & 14{,}541 & 237 & 66{,}411 \\
GraphHigher & 900 & 197 & 10{,}000 & 1{,}838 & 221 & 20{,}000 & 3{,}714 & 234 & 40{,}000 & 7{,}467 & 237 & 80{,}000 & 14{,}541 & 237 & 160{,}116 \\
GraphLower  & 7{,}505 & 237 & 160{,}000 & 11{,}258 & 237 & 80{,}000 & 13{,}134 & 237 & 40{,}000 & 14{,}072 & 237 & 20{,}000 & 14{,}541 & 237 & 10{,}116 \\
PS-CKGE  & 13{,}530 & 237 & 186{,}385 & 13{,}941 & 237 & 62{,}023 & 14{,}541 & 237 & 61{,}708 & -- & -- & -- & -- & -- & -- \\
\hline
\end{tabular}
}
\end{table*}
\label{sec:experiments}

In this section, we present an experimental evaluation designed to address the following research questions:
\begin{itemize}
    \item \textbf{RQ1: Evaluation bias.} How does taking into account entity interference change the performance reported by existing CKGE methods?
    \item \textbf{RQ2: Quantifying forgetting.} How effective are different CKGE learning methods at mitigating catastrophic forgetting when evaluating it with the corrected protocol?
    \item \textbf{RQ3: Decomposing forgetting.} How do representation drift and entity interference contribute to catastrophic forgetting across different KG evolution scenarios?
    \item \textbf{RQ4: Model sensitivity.} How do different KGE models behave in the continual learning setting, and how are they affected by catastrophic forgetting?
\end{itemize}

\subsection{Experimental Settings}
\label{sec:exp_settings}
\subsubsection{Datasets}
To observe the effect of the correct CKGE evaluation and study catastrophic forgetting, we use eight standard CKGE benchmark datasets with different characteristics (see Table~\ref{tab:datasets}). The first four datasets, ENTITY, RELATION, FACT, and HYBRID, were introduced in~\citep{LKGE} and consist of five snapshots. The datasets are derived from the FB15K-237 KG completion benchmark~\citep{fb15k} based on Freebase~\citep{freebase}. ENTITY, RELATION, and FACT are constructed such that each successive snapshot contains an increasingly larger number of entities, relations, and facts, respectively, while HYBRID increases all the three in a balanced way. This allows to assess how catastrophic forgetting behaves under different types of KG updates. The datasets GraphEqual, GraphHigher and GraphLower~\citep{incDE} also have five snapshots derived from FB15K-237, but they increase the number of triples between snapshots at different rates. This allows us to evaluate the effect of different update sizes in CKGE evaluation and catastrophic forgetting. Finally, the PS-CKGE dataset was introduced in~\citep{rethinking} as a new CKGE benchmark to assess pattern shifts that alter the graph structure.  

\subsubsection{Continual Learning Benchmark Methods}
We compare a range of continual learning techniques for CKGE, analyzing their behavior under the corrected evaluation proposed in Section~\ref{sec:limitations} and investigating the sources of catastrophic forgetting. As a first continual learning baseline, we take fine-tuning, which consists of training over the already learned embeddings with $\Delta\mathcal{T}_i$ using only the loss function $\mathcal{L}_{\mathcal{M}}$. For standard continual learning techniques, we  include the regularization method EWC~\citep{EWC}, which weights the importance of parameters using the Fisher Information Matrix (FIM), and the replay-based method EMR~\citep{EMR}, which mitigates forgetting by storing and replaying a subset of past triples during training. For CKGE approaches, we include those methods listed in Table~\ref{tab:ckge} for which the code is publicly available. Although these methods use various of the techniques presented in Section~\ref{sec:forgetting_mitigation}, they can be classified based on where they mostly intervene during training:
\paragraph{Loss Level}
\begin{itemize}
    \item LKGE~\citep{LKGE} uses a masked autoencoder for embedding reconstruction, embedding transfer, and regularization to mitigate catastrophic forgetting.
    \item FMR~\citep{FMR} mitigates catastrophic forgetting through a dual-level regularization strategy that adapts constraints based on learned knowledge, and rotates and aligns the parameter space to address FIM's limitations.
    \item ETT-CKGE~\citep{ETT-CKGE} introduces task-driven learnable tokens for efficient knowledge transfer between snapshots, reducing the need for explicit node scoring or graph traversal while preserving past information.
    \item DebiasedKGE~\citep{DebiasedKGE} mitigates knowledge interference and misalignment through disentangled representation learning and dual-view regularized alignment that constraints the magnitude and direction of embedding changes.
\end{itemize}
\paragraph{Model Level}
\begin{itemize}
    \item FastKGE~\citep{fastKGE} uses incremental low-rank adapters with adaptive rank allocation to more efficiently create embeddings for new entities and relations while isolating new knowledge to specific layers. 
    \item SAGE~\citep{SAGE}  adaptively expands embedding dimensions based on update scales and uses dynamic distillation to balance old and new knowledge.
\end{itemize}
\paragraph{Data Level}
\begin{itemize}
    \item incDE~\citep{incDE} uses hierarchical triple ordering and incremental distillation to respect KG structure and learning order, allowing new knowledge to be learned efficiently while old knowledge is preserved.
\end{itemize}

Finally, we also compute the best expected final performance by retraining the model with all triples $\mathcal{T}_n$ by using the loss function $\mathcal{L}_{\mathcal{M}}$.

\subsubsection{Metrics}
To evaluate CKGE predictive performance, we use standard Link Prediction metrics~\citep{KGAPP}. These are reported by aggregating the results of all test sets obtained after training with the final snapshot, weighting each test set by its number of triples:

\begin{itemize}
\item \textbf{Mean Reciprocal Rank (MRR)} indicates how early the correct entity appears in the ranked list by averaging the reciprocal of its rank across all incomplete triples.
\item \textbf{Hits@k} measures the proportion of incomplete triples for which the correct entity is ranked among the top-$k$ positions. In our experiments, we report Hits@1, Hits@3 and  Hits@10.
\end{itemize}

These metrics will be reported by also taking into account the interference from new entities, thus correcting CKGE evaluation as explained in Section~\ref{sec:limitations}.

Additionally, catastrophic forgetting is quantified by using the metric defined in Equation~\ref{eq:cf}, which will be derived from the corrected MRR evaluation metric.
\subsubsection{Settings}
For the CKGE methods, we adopt the hyperparameter configurations reported by the original authors~\citep{LKGE,incDE,fastKGE,FMR,ETT-CKGE,SAGE,DebiasedKGE}. For EWC and EMR, we use the same settings as in~\citep{LKGE}. To ensure fairness across models, we fix the embedding dimension to 200. For SAGE, which increases the embedding dimension over snapshots, we set the embedding size of the final snapshot to 200. For all the methods, TransE has been used as the underlying KGE model. To ensure reproducibility, we provide all datasets and implementation details in the accompanying repository.\footnote{https://github.com/gerardponsrecasens/RevisitingCKGE}
\subsection{Evaluation bias (RQ1)}
\label{sec:exp_correct}

\begin{table*}[t]
\caption{MRR Results across ENTITY, RELATION, FACT, and HYBRID Datasets with and without Accounting for Entity Interference.}
\label{tab:E-R-F-H}
\centering
\renewcommand{\arraystretch}{1.1}

\begin{tabularx}{\textwidth}{c
  *{3}{>{\centering\arraybackslash}X} |
  *{3}{>{\centering\arraybackslash}X} |
  *{3}{>{\centering\arraybackslash}X} |
  *{3}{>{\centering\arraybackslash}X}
}
\hline
\multirow{2}{*}{CKGE} &
\multicolumn{3}{c}{\textbf{ENTITY}} &
\multicolumn{3}{c}{\textbf{RELATION}} &
\multicolumn{3}{c}{\textbf{FACT}} &
\multicolumn{3}{c}{\textbf{HYBRID}} \\
\cline{2-13}
 &
$MRR^{\text{w/o}}$ & $MRR$ & \% & 
$MRR^{\text{w/o}}$ & $MRR$ & \% & 
$MRR^{\text{w/o}}$ & $MRR$ & \% & 
$MRR^{\text{w/o}}$ & $MRR$ & \%  \\
\hline

retraining &
0.237 & 0.211 & -11.0 &
0.220 & 0.220 & 0.0 &
0.205 & 0.204 & -0.5 &
0.227 & 0.210 & -7.5 \\

finetune &
0.163 & 0.143 & -12.3 &
0.089 & 0.089 & 0.0 &
0.173 & 0.172 & -0.6 &
0.138 & 0.126 & -8.7 \\

EWC &
0.228 & 0.185 & -18.9 &
0.157 & 0.157 & 0.0 &
0.201 & 0.199 & -1.0 &
0.187 & 0.173 & -7.5 \\

EMR &
0.169 & 0.148 & -12.4 &
0.109 & 0.109 & 0.0 &
0.174 & 0.173 & -0.6 &
0.139 & 0.128 & -7.9 \\

LKGE &
0.239 & 0.185 & -22.6 &
0.175 & 0.175 & 0.0 &
0.209 & 0.208 & -0.5 &
0.201 & 0.175 & -12.9 \\

FMR &
0.230 & 0.187 & -18.7 &
0.188 & 0.187 & -0.5 &
0.213 & 0.209 & -1.9 &
0.192 & 0.178 & -7.3 \\

ETT-CKGE &
0.259 & 0.200 & -22.8 &
0.195 & 0.195 & 0.0 &
0.214 & 0.212 & -0.9 &
0.223 & 0.192 & -13.9 \\

DebiasedKGE &
0.237 & 0.186 & -21.5 &
0.164 & 0.164 & 0.0 &
0.176 & 0.174 & -1.1 &
0.187 & 0.158 & -15.5 \\

FastKGE &
0.239 & 0.179 & -25.1 &
0.170 & 0.169 & -0.6 &
0.171 & 0.166 & -2.9 &
0.192 & 0.161 & -16.1 \\

SAGE &
0.249 & 0.193 & -22.5 &
0.180 & 0.180 & 0.0 &
0.217 & 0.215 & -0.9 &
0.203 & 0.176 & -13.3 \\

incDE &
0.253 & 0.186 & -26.5 &
0.198 & 0.198 & 0.0 &
0.216 & 0.213 & -1.4 &
0.224 & 0.185 & -17.4 \\

\hline
\end{tabularx}
\end{table*}

\begin{table*}[t]
\caption{MRR Results across GraphEqual, GraphHigher, GraphLower, and PS-CKGE Datasets with and without Accounting for Entity Interference.}
\label{tab:GQ-GH-GL-PS}
\centering
\renewcommand{\arraystretch}{1.05}

\begin{tabularx}{\textwidth}{c
  *{3}{>{\centering\arraybackslash}X} |
  *{3}{>{\centering\arraybackslash}X} |
  *{3}{>{\centering\arraybackslash}X} |
  *{3}{>{\centering\arraybackslash}X}
}
\hline
\multirow{2}{*}{CKGE} &
\multicolumn{3}{c}{\textbf{GraphEqual}} &
\multicolumn{3}{c}{\textbf{GraphHigher}} &
\multicolumn{3}{c}{\textbf{GraphLower}} &
\multicolumn{3}{c}{\textbf{PS-CKGE}} \\
\cline{2-13}
 &
$MRR^{\text{w/o}}$ & $MRR$ & \% &
$MRR^{\text{w/o}}$ & $MRR$ & \% &
$MRR^{\text{w/o}}$ & $MRR$ & \% &
$MRR^{\text{w/o}}$ & $MRR$ & \% \\
\hline

retraining &
0.222 & 0.202 & -9.0 &
0.230 & 0.207 & -10.0 &
0.215 & 0.206 & -4.2 &
0.214 & 0.214 & 0.0 \\

finetune &
0.180 & 0.164 & -8.9 &
0.199 & 0.180 & -9.5 &
0.187 & 0.181 & -3.2 &
0.145 & 0.145 & 0.0 \\

EWC &
0.206 & 0.185 & -10.2 &
0.197 & 0.165 & -16.2 &
0.211 & 0.200 & -5.2 &
0.202 & 0.201 & -0.5 \\

EMR &
0.184 & 0.167 & -9.2 &
0.204 & 0.186 & -8.8 &
0.183 & 0.175 & -4.4 &
0.146 & 0.146 & 0.0 \\

LKGE &
0.223 & 0.192 & -13.9 &
0.221 & 0.187 & -15.4 &
0.212 & 0.201 & -5.2 &
0.209 & 0.209 & 0.0 \\

FMR &
0.203 & 0.177 & -12.8 &
0.192 & 0.166 & -13.5 &
0.207 & 0.194 & -6.3 &
0.208 & 0.207 & -0.5 \\

ETT-CKGE &
0.228 & 0.190 & -16.7 &
0.228 & 0.192 & -15.8 &
0.203 & 0.191 & -5.9 &
0.188 & 0.187 & -0.5 \\

DebiasedKGE &
0.217 & 0.182 & -16.1 &
0.205 & 0.168 & -18.0 &
0.199 & 0.181 & -9.0 &
0.191 & 0.191 & 0.0 \\

FastKGE &
0.214 & 0.170 & -20.6 &
0.194 & 0.159 & -18.0 &
0.215 & 0.184 & -14.4 &
0.204 & 0.202 & -1.0 \\

SAGE &
0.228 & 0.186 & -18.4 &
0.231 & 0.196 & -15.2 &
0.210 & 0.196 & -6.7 &
0.214 & 0.214 & 0.0 \\

incDE &
0.234 & 0.185 & -20.9 &
0.228 & 0.187 & -18.0 &
0.226 & 0.207 & -8.4 &
0.220 & 0.219 & -0.5 \\

\hline
\end{tabularx}
\end{table*}

Tables~\ref{tab:E-R-F-H}-\ref{tab:GQ-GH-GL-PS}, present the evaluation results for MRR for all datasets and CKGE methods. The results for Hits@k metrics can be found in Appendix~\ref{ap:a}. The evaluation is reported without entity interference ($MRR^{w/o}$) as done in previous works, and also the corrected version ($MRR$) where interference is considered. As discussed in Section~\ref{sec:quantify}, incorporating entity interference leads to $MRR^{w/o} \geq MRR$. 

A first observation is that, as expected, the number of new entities introduced across snapshots (see Table~\ref{tab:datasets}) plays an important role. For  datasets with smaller growth in the number of new entities (e.g., PS-CKGE and RELATION, where the total growth from the first to the last snapshot is 7\% and 26\%, respectively), the differences in evaluation when incorporating entity interference are minor. In contrast, for datasets with a large number of new entities (e.g., GraphHigher, GraphEqual and ENTITY, where the growth is  1516\%, 400\% and 400\%, respectively), the differences are substantial,  with some CKGE methods losing more than 25\% of their originally reported performance. 

Importantly, for these datasets with many new entities, retraining is outperformed by several CKGE methods when not considering entity interference. For instance, in the ENTITY dataset, retraining is outperformed by 5 out of 7 CKGE methods. However, this is no longer the case after correcting the evaluation, as retraining becomes then the best performing method. Concretely, when retraining there is an 11\% performance drop when accounting  for interference, whereas the methods previously outperforming it experience reductions of more than 22\%. Therefore, the representations learned when training the model with all the data at once are more robust to interferences, suggesting that future CKGE methods could benefit from catastrophic forgetting mitigation strategies that explicitly address interference.

\subsection{Quantifying Catastrophic Forgetting (RQ2)}
\label{sec:exp_decompose}

\begin{table*}[t]
\caption{Catastrophic Forgetting Derived from the Corrected Evaluation Values for Different Datasets .}
\label{tab:ep_cf}
\centering
\renewcommand{\arraystretch}{1.1}

\begin{tabularx}{\textwidth}{
  l
  *{9}{>{\centering\arraybackslash}X}
}
\hline
\textbf{CKGE} &
\textbf{ENTITY} &
\textbf{RELATION} &
\textbf{FACT} &
\textbf{HYBRID} &
\textbf{GEqual} &
\textbf{GHigher} &
\textbf{GLower} &
\textbf{PS} &
\textbf{Average} \\
\hline
finetune     & -0.260 & -0.220 & -0.121 & -0.251 & -0.103 & -0.218 & -0.033 & -0.116 & -0.165\\
EWC          & -0.110 & -0.011 & -0.024 & -0.086 & -0.051 & -0.212 & -0.017 & -0.008 & \textbf{-0.065}\\
EMR          & -0.245 & -0.179 & -0.120 & -0.232 & -0.099 & -0.202 & -0.045 & -0.114 & -0.154\\
LKGE         & -0.122 & -0.078 & -0.054 & -0.110 & -0.051 & -0.149 & -0.021 & -0.017 & -0.075\\
FMR          & -0.101 & -0.024 & -0.041 & -0.098 & -0.067 & -0.231 & -0.030 & -0.021 & -0.077\\
ETT-CKGE     & -0.133 & -0.053 & -0.035 & -0.065 & -0.080 & -0.170 & -0.045 & -0.071 & -0.082\\
DebiasedKGE  & -0.113 & -0.072 & -0.075 & -0.115 & -0.065 & -0.203 & -0.047 & -0.033 & -0.090\\
FastKGE      & -0.138 & -0.002 & -0.011 & -0.054 & -0.084 & -0.202 & -0.040 & -0.003 & -0.067\\
SAGE         & -0.142 & -0.098 & -0.056 & -0.128 & -0.081 & -0.161 & -0.036 & -0.028 & -0.091\\
incDE        & -0.174 & -0.044 & -0.025 & -0.096 & -0.087 & -0.180 & -0.026 & -0.020 & -0.082\\
\midrule
Average      & -0.154 & -0.078 & -0.056 & -0.124 & -0.077 & -0.193 & \textbf{-0.034} & -0.043 & -- \\
\hline
\end{tabularx}

\end{table*}

Table~\ref{tab:ep_cf} reports catastrophic forgetting  across continual learning methods and datasets. The metric corresponds to the catastrophic forgetting measure defined in Equation~\ref{eq:cf} and it is derived from the corrected evaluation metrics (i.e., taking entity interference into account). 

Overall, as expected, non-regularized methods (i.e., finetune and EMR) experience the higher amounts of catastrophic forgetting. This is particularly pronounced in datasets with a smaller proportion of new entities. For instance, on the PS-CKGE dataset, the amount of forgetting observed for non-regularized methods is up to two orders of magnitude higher that that of highly regularized methods such as EWC or FastKGE.

Methods such as EWC, LKGE or SAGE were reported in~\citep{SAGE} to achieve near-optimal resistance to catastrophic forgetting. However, this conclusion no longer holds once entity interference is taken into account during the evaluation. For instance, this can be observed in the ENTITY and GraphHigher datasets in Table~\ref{tab:ep_cf}.

\subsection{Decomposing Catastrophic Forgetting (RQ3)}

In order to better observe the two causes of forgetting presented in Section~\ref{sec:understanding}, we focus on the test set of Snapshot 0 as the embeddings of entities and relations present in it undergo the largest number of continual updates. Table~\ref{tab:sources} reports the test MRR of Snaphshot 0 at different stages. Specifically, we first report the initial performance $MRR_{0,0}$, followed by the performance after completing all continual learning steps, both without accounting for entity interference ($\mathrm{MRR}^{\text{w/o}}_{N,0}$) and with the corrected evaluation ($MRR_{N,0}$). The results are reported for two datasets with a large number of new entities (ENTITY and GraphHigher) and for a dataset with a smaller growth in entities (RELATION). 

It can be observed that for ENTITY and GraphHigher datasets, the effect of correcting the evaluation by including entity interference is highly noticeable, with a decrease in reported performance which can be higher than 50\%. Remarkably, in some scenarios the uncorrected evaluation suggests that performance remains stable or even improves over time, whereas the corrected evaluation reveals substantial forgetting. For instance, for the methods EWC, LKGE, FMR and DebiasedKGE on the GraphHigher dataset, $\mathrm{MRR}^{\text{w/o}}_{N,0}$ increases relative to $MRR_{0,0}$, while the corrected metric $MRR_{N,0}$ actually decreases by 40-60\%. Not including entity interference can be particularly misleading for methods such as FastKGE, where the embedding parameters are frozen. In this case, under the standard evaluation protocol, performance remains constant, giving the impression that no catastrophic forgetting occurs~\citep{DyGM}. 

For the RELATION dataset, which has fewer new entities, the effect can be considered negligible and representation drift is the primary source of forgetting.

\begin{table*}[t]
\centering
\caption{Breaking Down the Sources of Catastrophic Forgetting}
\renewcommand{\arraystretch}{1.1}

\begin{tabularx}{\textwidth}{c
  *{3}{>{\centering\arraybackslash}X} |
  *{3}{>{\centering\arraybackslash}X} |
  *{3}{>{\centering\arraybackslash}X}
}
\hline
\multirow{2}{*}{CKGE} &
\multicolumn{3}{c}{\textbf{ENTITY}} &
\multicolumn{3}{c}{\textbf{GraphHigher}} &
\multicolumn{3}{c}{\textbf{RELATION}} \\
\cline{2-10}
 & 
$MRR_{0,0}$ & $\mathrm{MRR}^{\text{w/o}}_{N,0}$ & $MRR_{N,0}$ &
$MRR_{0,0}$ & $\mathrm{MRR}^{\text{w/o}}_{N,0}$ & $MRR_{N,0}$&
$MRR_{0,0}$ & $\mathrm{MRR}^{\text{w/o}}_{N,0}$ & $MRR_{N,0}$ \\
\hline

finetune &
0.303 & 0.180 & 0.130 &
0.372 & 0.328 & 0.191 &
0.259 & 0.069 & 0.069 \\

EWC &
0.303 & 0.304 & 0.192 &
0.372 & 0.383 & 0.181 &
0.259 & 0.241 & 0.241 \\

EMR &
0.303 & 0.190 & 0.139 &
0.372 & 0.358 & 0.219 &
0.259 & 0.109 & 0.108 \\

LKGE &
0.312 & 0.303 & 0.175 &
0.394 & 0.403 & 0.194 &
0.263 & 0.192 & 0.191 \\

FMR &
0.310 & 0.310 & 0.191 &
0.390 & 0.391 & 0.228 &
0.264 & 0.262 & 0.262 \\

ETT-CKGE &
0.336 & 0.326 & 0.186 &
0.439 & 0.437 & 0.207 &
0.276 & 0.193 & 0.192 \\

DebiasedKGE &
0.312 & 0.310 & 0.176 &
0.375 & 0.377 & 0.169 &
0.262 & 0.186 & 0.186 \\

FastKGE &
0.332 & 0.332 & 0.179 &
0.433 & 0.433 & 0.185 &
0.278 & 0.278 & 0.276 \\

SAGE &
0.335 & 0.308 & 0.170 &
0.437 & 0.432 & 0.206 &
0.270 & 0.180 & 0.179 \\

incDE &
0.338 & 0.308 & 0.162 &
0.434 & 0.425 & 0.167 &
0.277 & 0.214 & 0.213 \\

\hline
\end{tabularx}

\label{tab:sources}
\end{table*}

\subsection{Catastrophic Forgetting in KGE Models (RQ4)}
\label{sec:models}

 All CKGE models used in the previous experiments rely on TransE~\citep{TransE} as their underlying embedding model. TransE is a foundational translational KGE model which views the relations as translation by addition from the head entity embedding to the tail entity embedding. Several extensions of this translational idea have been proposed, such as TransH~\citep{TransH} or TransR~\citep{TransR}, which generalize the approach by introducing relation-specific hyperplanes or different embedding spaces to better capture heterogeneous relation types. Models like RotatE~\citep{RotatE} and TorusE~\citep{toruse} introduce rotations or translations in manifolds, enabling the modeling of complex relation patterns such as symmetry and inversion. Beyond translational approaches, there also exist semantic models, which define scoring functions based on multiplicative interactions between entity and relation embeddings. DistMult~\citep{DistMult} and ComplEx~\citep{ComplEx} represent the simplest forms of semantic models, by using real and complex-valued embeddings, respectively. QuatE~\citep{QuatE} extends ComplEx by utilizing quaternion algebra, offering a richer way to capture multi-relational patterns. SimplE~\citep{SimplE} also uses bilinear scoring but optimizes for simpler and more intuitive representations. These models are particularly effective for capturing complex relational patterns such as symmetry, inversion, and composition.

 In this experiment, we investigate how the choice of a KGE model affects the CKGE learning process, with a particular focus on catastrophic forgetting. We use the ENTITY dataset as it allows us to observe both representation drift and entity interference, and finetuning as the continual learning strategy. The latter choice is motivated by the fact that most CKGE methods are implemented exclusively for TransE and are not applicable to other KGE models. First, we obtain the best initial embeddings possible for Snapshot 0 by grid search over the embedding dimension $d\in[50,100,150,200]$ and learning rate $lr\in[10^{-2}, 10^{-3}, 10^{-4}, 10^{-5}]$. Starting from these embeddings, we continue training in a continual learning setting for the remaining snapshots, again performing a grid search over incremental learning rates centered around the optimal rate identified for the initial snapshot: $lr_{inc}\in[10^{-1}\cdot lr,lr,10\cdot lr]$.

 The results for the experiments are summarized in Table~\ref{tab:kge_models}, which reports the final MRR, the catastrophic forgetting metric, the performance on Snapshot 0 before and after continual learning ($MRR_{0,0}$ and $MRR_{N,0}$) with its uncorrected evaluation without entity interference ($\mathrm{MRR}^{\text{w/o}}_{N,0}$), and $\Omega_{new}$, which is a metric which averages the knowledge learned when an incremental snapshot is introduced (i.e., $MRR_{i,i} \text{ }\forall \text{ }i>0$).

 First of all, it can be observed that a higher initial performance (i.e., $MRR_{0,0}$) does not always translate to more robust continual updates (see for instance BoxE and QuatE or SimplE and TransR). Among all the approaches, RotatE and BoxE achieve the highest final MRR values, yet they suffer from considerable catastrophic forgetting. For instance, BoxE performance over the first snapshot gets reduced by more than 50\% due to a combination of both representation drift (see $MRR_{0,0}$ vs $MRR_{N,0}^{w/o}$) and entity interference (see $MRR_{N,0}^{w/o}$ vs. $MRR_{N,0}$).

 In contrast, TransR stands out as the most stable model, as it exhibits the lowest catastrophic forgetting and only marginal impact from entity interference. Other translational models, such as TransE and TransH, also show moderate forgetting and are less affected by entity interference when compared to more expressive alternatives. 

 Semantic models are the most consistently affected by continual learning. These models show  substantial catastrophic forgetting, both by representation drift and entity interference. 

 When considering $\Omega_{new}$, we can observe one of the trade-offs from continual learning, the plasticity versus stability~\citep{plasticty-stability}. Models like QuatE, which rapidly adapt to new information, have high levels of forgetting, while more stable models such as the translational ones, fall behind in new knowledge acquisition. 

 In summary, semantic models are the ones more affected by both representation drift and entity interference, yet they are better at knowledge acquisition. Translational models, and TransR in particular, show great stability and robustness to interference, but they fall behind in knowledge acquisition. 

\begin{table*}[t]
\centering
\caption{Overall Performance and Catastrophic Forgetting Analysis across Different KGE Models in the ENTITY Dataset.}
\label{tab:kge_models}
\begin{tabular}{llcccccc}
\toprule
KGE Type & Model & MRR & CF & $MRR_{0,0}$ & $\mathrm{MRR}^{\text{w/o}}_{N,0}$ & $MRR_{N,0}$ & $\Omega_{new}$ \\
\midrule
\multirow{6}{*}{Translational}
 & TransE  & 0.226  & -0.150 & 0.339 & 0.293 & 0.215 & 0.298 \\
 & TransH  & 0.177 & -0.183 & 0.374 & 0.209 & 0.156 & 0.251 \\
 & TransR  & 0.188 & \textbf{-0.028} & 0.295 & 0.273 & \textbf{0.270} & 0.183 \\
 & RotatE & \textbf{0.295} & -0.192 & 0.407 & \textbf{0.344} & 0.223 & 0.442 \\
 & TorusE & 0.231 & -0.197 & 0.398 & 0.176 & 0.072 & 0.378 \\
 & BoxE   & 0.258 & -0.234 & \textbf{0.488} & 0.337 & 0.205 & 0.430 \\
\midrule
\multirow{4}{*}{Semantic}
 & DistMult & 0.170 & -0.276 & 0.359 & 0.210 & 0.119 & 0.315 \\
 & ComplEx & 0.112 & -0.458 & 0.354 & 0.081 & 0.058 & 0.378 \\
 & QuatE   & 0.159 & -0.382 & 0.466 & 0.086 & 0.049 & \textbf{0.451} \\
 & SimplE  & 0.097 & -0.427 & 0.297 & 0.097 & 0.039 & 0.281 \\
\bottomrule
\end{tabular}
\end{table*}

\section{Conclusions}
\label{sec:conclusions}

In this work, we present an in-depth analysis of catastrophic forgetting in CKGE, demonstrating that the existing view is incomplete, as it focuses on representation drift of previously learned embeddings while overlooking the impact of newly introduced entities. We identify \textit{entity interference} as a second source of forgetting, demonstrating that even perfectly preserved embeddings of old entities can yield degraded predictions once new entities are incorporated to the embedding space.

Based on these observations, we showed that the standard CKGE evaluation protocol systematically excludes entity interference by restricting candidate entities to those present at the time each snapshot is learned. Through extensive experiments on widely used benchmarks, we demonstrate that this leads to an overestimation of CKGE performance and, consequently, to misleading conclusions about the effectiveness of existing CKGE methods. This issue is particularly critical when assessing catastrophic forgetting, as approaches that may appear robust under the standard evaluation in reality suffer from performance degradation once interference from new entities is properly considered. To address this, we propose a metric for properly assessing catastrophic forgetting in the CKGE setting, and analyze the contribution of the different sources of forgetting. Our results reveal that entity interference is the dominant source of forgetting in scenarios with substantial growth in the number of entities, whereas representation drift remains the primary factor when entity growth is limited. Moreover, retraining on the full KG consistently yields embeddings that are more robust to interference, highlighting a gap between the learning process of current CKGE methods and the ideal continual learning behavior they aim to approximate. 

Finally, we analyze the behavior of different KGE models in the continual learning setting, finding that translational models are more stable under both sources of forgetting, while semantic models adapt better to new knowledge in the expense of experiencing higher levels of forgetting.

Overall, our findings highlight the need to revisit both evaluation protocols and catastrophic forgetting mitigation strategies in CKGE. In particular, future work should explicitly account for entity interference, for example by designing learning objectives or architectures that are more robust to interference, or by exploring alternative ways to handling the growing candidate entity set during inference. We believe that the proposed evaluation protocol and forgetting metric will enable more reliable assessment and comparison, supporting further progress in the development of CKGE methods.


\printcredits

\section*{Acknowledgments}
This work is supported by the Horizon Europe Programme under
GA.101135513 (CyclOps) and the Spanish Ministerio de Ciencia e
Innovación under project PID2023-152841OA-I00/AEI/10.13039/501100011033 (TALC). Anna Queralt is a Serra Húnter Fellow.
\bibliographystyle{cas-model2-names}

\bibliography{references}

\clearpage
\appendix

\section{Additional Results}
\label{ap:a}
In this section, the corrected evaluation results are shown for the metrics Hits@1 (Table~\ref{tab:E-R-F-H-hits1} and Table~\ref{tab:GQ-GH-GL-PS-hits1}), Hits@3 (Table~\ref{tab:E-R-F-H-hits3} and Table~\ref{tab:GQ-GH-GL-PS-hits3}) and Hits@10 (Table~\ref{tab:E-R-F-H-hits10} and Table~\ref{tab:GQ-GH-GL-PS-hits10}).

\clearpage

\begin{table}[B]
\caption{Hits@1 Results across ENTITY, RELATION, FACT, and HYBRID Datasets with and without Accounting for Entity Interference.}
\label{tab:E-R-F-H-hits1}
\centering
\renewcommand{\arraystretch}{1.1}

\begin{tabularx}{\textwidth}{c
  *{3}{>{\centering\arraybackslash}X} |
  *{3}{>{\centering\arraybackslash}X} |
  *{3}{>{\centering\arraybackslash}X} |
  *{3}{>{\centering\arraybackslash}X}
}
\hline
\multirow{2}{*}{CKGE} &
\multicolumn{3}{c}{\textbf{ENTITY}} &
\multicolumn{3}{c}{\textbf{RELATION}} &
\multicolumn{3}{c}{\textbf{FACT}} &
\multicolumn{3}{c}{\textbf{HYBRID}} \\
\cline{2-13}
 &
$H@1^{w/o}$ & $H@1$ & \% & 
$H@1^{w/o}$ & $H@1$ & \% & 
$H@1^{w/o}$ & $H@1$ & \% & 
$H@1^{w/o}$ & $H@1$ & \%  \\
\hline

retraining &
0.136 & 0.123 & -9.6 &
0.128 & 0.128 & 0.0 &
0.117 & 0.116 & -0.9 &
0.134 & 0.121 & -9.7 \\

finetune &
0.083 & 0.075 & -9.6 &
0.038 & 0.038 & 0.0 &
0.090 & 0.090 & 0.0 &
0.072 & 0.063 & -12.5 \\

EWC &
0.129 & 0.107 & -17.1 &
0.086 & 0.086 & 0.0 &
0.112 & 0.111 & -0.9 &
0.103 & 0.096 & -6.8 \\

EMR &
0.088 & 0.078 & -11.4 &
0.051 & 0.051 & 0.0 &
0.092 & 0.092 & 0.0 &
0.073 & 0.066 & -9.6 \\

LKGE &
0.140 & 0.109 & -22.1 &
0.092 & 0.092 & 0.0 &
0.121 & 0.120 & -0.8 &
0.114 & 0.095 & -16.7 \\

FMR &
0.134 & 0.112 & -16.4 &
0.098 & 0.098 & 0.0 &
0.126 & 0.125 & -0.8 &
0.109 & 0.100 & -8.3 \\

ETT-CKGE &
0.155 & 0.118 & -23.9 &
0.107 & 0.107 & 0.0 &
0.123 & 0.122 & -0.8 &
0.131 & 0.106 & -19.1 \\

DebiasedKGE &
0.139 & 0.112 & -19.4 &
0.086 & 0.086 & 0.0 &
0.085 & 0.084 & -1.2 &
0.103 & 0.082 & -20.4 \\

FastKGE &
0.144 & 0.109 & -24.3 &
0.100 & 0.099 & -1.0 &
0.106 & 0.104 & -1.9 &
0.114 & 0.091 & -20.2 \\

SAGE &
0.150 & 0.117 & -22.0 &
0.100 & 0.099 & -1.0 &
0.129 & 0.128 & -0.8 &
0.115 & 0.094 & -18.3 \\

incDE &
0.150 & 0.109 & -27.3 &
0.109 & 0.109 & 0.0 &
0.127 & 0.126 & -0.8 &
0.132 & 0.102 & -22.7 \\

\hline
\end{tabularx}
\end{table}

\begin{table}[]
\caption{Hits@1 Results across GraphEqual, GraphHigher, GraphLower, and PS-CKGE Datasets with and without Accounting for Entity Interference.}
\label{tab:GQ-GH-GL-PS-hits1}
\centering
\renewcommand{\arraystretch}{1.05}

\begin{tabularx}{\textwidth}{c
  *{3}{>{\centering\arraybackslash}X} |
  *{3}{>{\centering\arraybackslash}X} |
  *{3}{>{\centering\arraybackslash}X} |
  *{3}{>{\centering\arraybackslash}X}
}
\hline
\multirow{2}{*}{CKGE} &
\multicolumn{3}{c}{\textbf{GraphEqual}} &
\multicolumn{3}{c}{\textbf{GraphHigher}} &
\multicolumn{3}{c}{\textbf{GraphLower}} &
\multicolumn{3}{c}{\textbf{PS-CKGE}} \\
\cline{2-13}
 &
$H@1^{w/o}$ & $H@1$ & \% &
$H@1^{w/o}$ & $H@1$ & \% &
$H@1^{w/o}$ & $H@1$ & \% &
$H@1^{w/o}$ & $H@1$ & \% \\
\hline

retraining &
0.125 & 0.114 & -8.8 &
0.130 & 0.119 & -8.5 &
0.122 & 0.117 & -4.1 &
0.126 & 0.125 & -0.8 \\

finetune &
0.094 & 0.085 & -9.6 &
0.110 & 0.099 & -10.0 &
0.100 & 0.096 & -4.0 &
0.075 & 0.075 & 0.0 \\

EWC &
0.111 & 0.101 & -9.0 &
0.106 & 0.090 & -15.1 &
0.117 & 0.112 & -4.3 &
0.117 & 0.116 & -0.9 \\

EMR &
0.098 & 0.090 & -8.2 &
0.115 & 0.105 & -8.7 &
0.097 & 0.093 & -4.1 &
0.078 & 0.077 & -1.3 \\

LKGE &
0.126 & 0.109 & -13.5 &
0.129 & 0.110 & -14.7 &
0.118 & 0.112 & -5.1 &
0.122 & 0.121 & -0.8 \\

FMR &
0.113 & 0.099 & -12.4 &
0.109 & 0.095 & -12.8 &
0.114 & 0.108 & -5.3 &
0.122 & 0.121 & -0.8 \\

ETT-CKGE &
0.128 & 0.106 & -17.2 &
0.133 & 0.113 & -15.0 &
0.111 & 0.105 & -5.4 &
0.107 & 0.106 & -0.9 \\

DebiasedKGE &
0.119 & 0.102 & -14.3 &
0.116 & 0.097 & -16.4 &
0.106 & 0.099 & -6.6 &
0.104 & 0.104 & 0.0 \\

FastKGE &
0.124 & 0.100 & -19.4 &
0.114 & 0.094 & -17.5 &
0.123 & 0.107 & -13.0 &
0.125 & 0.124 & -0.8 \\

SAGE &
0.130 & 0.105 & -19.2 &
0.138 & 0.117 & -15.2 &
0.117 & 0.110 & -6.0 &
0.128 & 0.128 & 0.0 \\

incDE &
0.133 & 0.103 & -22.6 &
0.135 & 0.111 & -17.8 &
0.129 & 0.119 & -7.8 &
0.132 & 0.131 & -0.8 \\

\hline
\end{tabularx}
\end{table}

\begin{table}
\caption{Hits@3 Results across ENTITY, RELATION, FACT, and HYBRID Datasets with and without Accounting for Entity Interference.}
\label{tab:E-R-F-H-hits3}
\centering
\renewcommand{\arraystretch}{1.1}

\begin{tabularx}{\textwidth}{c
  *{3}{>{\centering\arraybackslash}X} |
  *{3}{>{\centering\arraybackslash}X} |
  *{3}{>{\centering\arraybackslash}X} |
  *{3}{>{\centering\arraybackslash}X}
}
\hline
\multirow{2}{*}{CKGE} &
\multicolumn{3}{c}{\textbf{ENTITY}} &
\multicolumn{3}{c}{\textbf{RELATION}} &
\multicolumn{3}{c}{\textbf{FACT}} &
\multicolumn{3}{c}{\textbf{HYBRID}} \\
\cline{2-13}
 &
$H@3^{w/o}$ & $H@3$ & \% & 
$H@3^{w/o}$ & $H@3$ & \% & 
$H@3^{w/o}$ & $H@3$ & \% & 
$H@3^{w/o}$ & $H@3$ & \%  \\
\hline

retraining &
0.275 & 0.239 & -13.1 &
0.252 & 0.252 & 0.0 &
0.232 & 0.230 & -0.9 &
0.260 & 0.238 & -8.5 \\

finetune &
0.185 & 0.160 & -13.5 &
0.100 & 0.100 & 0.0 &
0.194 & 0.193 & -0.5 &
0.154 & 0.141 & -8.4 \\

EWC &
0.264 & 0.207 & -21.6 &
0.182 & 0.181 & -0.5 &
0.229 & 0.227 & -0.9 &
0.218 & 0.199 & -8.7 \\

EMR &
0.193 & 0.166 & -14.0 &
0.123 & 0.123 & 0.0 &
0.195 & 0.193 & -1.0 &
0.154 & 0.142 & -7.8 \\

LKGE &
0.273 & 0.205 & -24.9 &
0.202 & 0.202 & 0.0 &
0.236 & 0.234 & -0.8 &
0.228 & 0.197 & -13.6 \\

FMR &
0.262 & 0.207 & -21.0 &
0.220 & 0.220 & 0.0 &
0.240 & 0.236 & -1.7 &
0.217 & 0.199 & -8.3 \\

ETT-CKGE &
0.303 & 0.225 & -25.7 &
0.230 & 0.229 & -0.4 &
0.247 & 0.243 & -1.6 &
0.260 & 0.221 & -15.0 \\

DebiasedKGE &
0.271 & 0.205 & -24.4 &
0.186 & 0.186 & 0.0 &
0.208 & 0.206 & -1.0 &
0.214 & 0.178 & -16.8 \\

FastKGE &
0.273 & 0.198 & -27.5 &
0.200 & 0.199 & -0.5 &
0.195 & 0.189 & -3.1 &
0.224 & 0.185 & -17.4 \\

SAGE &
0.288 & 0.217 & -24.7 &
0.209 & 0.209 & 0.0 &
0.247 & 0.245 & -0.8 &
0.235 & 0.203 & -13.6 \\

incDE &
0.294 & 0.209 & -28.9 &
0.235 & 0.235 & 0.0 &
0.250 & 0.246 & -1.6 &
0.262 & 0.212 & -19.1 \\

\hline
\end{tabularx}
\end{table}

\begin{table*}[h]
\caption{Hits@3 Results across GraphEqual, GraphHigher, GraphLower, and PS-CKGE Datasets with and without Accounting for Entity Interference.}
\label{tab:GQ-GH-GL-PS-hits3}
\centering
\renewcommand{\arraystretch}{1.05}

\begin{tabularx}{\textwidth}{c
  *{3}{>{\centering\arraybackslash}X} |
  *{3}{>{\centering\arraybackslash}X} |
  *{3}{>{\centering\arraybackslash}X} |
  *{3}{>{\centering\arraybackslash}X}
}
\hline
\multirow{2}{*}{CKGE} &
\multicolumn{3}{c}{\textbf{GraphEqual}} &
\multicolumn{3}{c}{\textbf{GraphHigher}} &
\multicolumn{3}{c}{\textbf{GraphLower}} &
\multicolumn{3}{c}{\textbf{PS-CKGE}} \\
\cline{2-13}
 &
$H@3^{w/o}$ & $H@3$ & \% &
$H@3^{w/o}$ & $H@3$ & \% &
$H@3^{w/o}$ & $H@3$ & \% &
$H@3^{w/o}$ & $H@3$ & \% \\
\hline

retraining &
0.254 & 0.226 & -11.0 &
0.264 & 0.233 & -11.7 &
0.244 & 0.231 & -5.3 &
0.241 & 0.241 & 0.0 \\

finetune &
0.204 & 0.182 & -10.8 &
0.226 & 0.202 & -10.6 &
0.212 & 0.204 & -3.8 &
0.162 & 0.162 & 0.0 \\

EWC &
0.234 & 0.207 & -11.5 &
0.225 & 0.183 & -18.7 &
0.240 & 0.226 & -5.8 &
0.228 & 0.227 & -0.4 \\

EMR &
0.208 & 0.186 & -10.6 &
0.230 & 0.207 & -10.0 &
0.204 & 0.195 & -4.4 &
0.162 & 0.162 & 0.0 \\

LKGE &
0.254 & 0.215 & -15.4 &
0.251 & 0.209 & -16.7 &
0.241 & 0.227 & -5.8 &
0.237 & 0.237 & 0.0 \\

FMR &
0.229 & 0.198 & -13.5 &
0.214 & 0.182 & -15.0 &
0.234 & 0.218 & -6.8 &
0.234 & 0.233 & -0.4 \\

ETT-CKGE &
0.264 & 0.215 & -18.6 &
0.260 & 0.213 & -18.1 &
0.232 & 0.217 & -6.5 &
0.212 & 0.211 & -0.5 \\

DebiasedKGE &
0.249 & 0.204 & -18.1 &
0.233 & 0.183 & -21.5 &
0.226 & 0.204 & -9.7 &
0.219 & 0.218 & -0.5 \\

FastKGE &
0.243 & 0.187 & -23.0 &
0.215 & 0.169 & -21.4 &
0.247 & 0.209 & -15.4 &
0.232 & 0.230 & -0.9 \\

SAGE &
0.264 & 0.209 & -20.8 &
0.266 & 0.221 & -16.9 &
0.241 & 0.222 & -7.9 &
0.244 & 0.244 & 0.0 \\

incDE &
0.271 & 0.207 & -23.6 &
0.261 & 0.207 & -20.7 &
0.260 & 0.235 & -9.6 &
0.251 & 0.250 & -0.4 \\

\hline
\end{tabularx}
\end{table*}

\begin{table*}[h]
\caption{Hits@10 Results across ENTITY, RELATION, FACT, and HYBRID Datasets with and without Accounting for Entity Interference.}
\label{tab:E-R-F-H-hits10}
\centering
\renewcommand{\arraystretch}{1.1}

\begin{tabularx}{\textwidth}{c
  *{3}{>{\centering\arraybackslash}X} |
  *{3}{>{\centering\arraybackslash}X} |
  *{3}{>{\centering\arraybackslash}X} |
  *{3}{>{\centering\arraybackslash}X}
}
\hline
\multirow{2}{*}{CKGE} &
\multicolumn{3}{c}{\textbf{ENTITY}} &
\multicolumn{3}{c}{\textbf{RELATION}} &
\multicolumn{3}{c}{\textbf{FACT}} &
\multicolumn{3}{c}{\textbf{HYBRID}} \\
\cline{2-13}
 &
$H@10^{w/o}$ & $H@10$ & \% & 
$H@10^{w/o}$ & $H@10$ & \% & 
$H@10^{w/o}$ & $H@10$ & \% & 
$H@10^{w/o}$ & $H@10$ & \%  \\
\hline

retraining &
0.434 & 0.390 & -10.1 &
0.405 & 0.405 & 0.0 &
0.385 & 0.382 & -0.8 &
0.415 & 0.392 & -5.5 \\

finetune &
0.318 & 0.282 & -11.3 &
0.186 & 0.186 & 0.0 &
0.344 & 0.341 & -0.9 &
0.264 & 0.248 & -6.1 \\

EWC &
0.422 & 0.340 & -19.4 &
0.297 & 0.296 & -0.3 &
0.382 & 0.378 & -1.0 &
0.351 & 0.326 & -7.1 \\

EMR &
0.327 & 0.289 & -11.6 &
0.221 & 0.221 & 0.0 &
0.344 & 0.341 & -0.9 &
0.264 & 0.248 & -6.1 \\

LKGE &
0.433 & 0.338 & -21.9 &
0.338 & 0.338 & 0.0 &
0.391 & 0.387 & -1.0 &
0.373 & 0.335 & -10.2 \\

FMR &
0.419 & 0.336 & -19.8 &
0.363 & 0.363 & 0.0 &
0.387 & 0.378 & -2.3 &
0.357 & 0.331 & -7.3 \\

ETT-CKGE &
0.460 & 0.360 & -21.7 &
0.362 & 0.361 & -0.3 &
0.396 & 0.390 & -1.5 &
0.402 & 0.358 & -10.9 \\

DebiasedKGE &
0.432 & 0.336 & -22.2 &
0.322 & 0.322 & 0.0 &
0.355 & 0.351 & -1.1 &
0.355 & 0.312 & -12.1 \\

FastKGE &
0.420 & 0.314 & -25.2 &
0.297 & 0.295 & -0.7 &
0.290 & 0.279 & -3.8 &
0.342 & 0.297 & -13.2 \\

SAGE &
0.436 & 0.343 & -21.3 &
0.336 & 0.336 & 0.0 &
0.394 & 0.390 & -1.0 &
0.370 & 0.335 & -9.5 \\

incDE &
0.446 & 0.335 & -24.9 &
0.370 & 0.370 & 0.0 &
0.392 & 0.385 & -1.8 &
0.402 & 0.349 & -13.2 \\

\hline
\end{tabularx}
\end{table*}

\begin{table*}[h]
\caption{Hits@10 Results across GraphEqual, GraphHigher, GraphLower, and PS-CKGE Datasets with and without Accounting for Entity Interference.}
\label{tab:GQ-GH-GL-PS-hits10}
\centering
\renewcommand{\arraystretch}{1.05}

\begin{tabularx}{\textwidth}{c
  *{3}{>{\centering\arraybackslash}X} |
  *{3}{>{\centering\arraybackslash}X} |
  *{3}{>{\centering\arraybackslash}X} |
  *{3}{>{\centering\arraybackslash}X}
}
\hline
\multirow{2}{*}{CKGE} &
\multicolumn{3}{c}{\textbf{GraphEqual}} &
\multicolumn{3}{c}{\textbf{GraphHigher}} &
\multicolumn{3}{c}{\textbf{GraphLower}} &
\multicolumn{3}{c}{\textbf{PS-CKGE}} \\
\cline{2-13}
 &
$H@10^{w/o}$ & $H@10$ & \% &
$H@10^{w/o}$ & $H@10$ & \% &
$H@10^{w/o}$ & $H@10$ & \% &
$H@10^{w/o}$ & $H@10$ & \% \\
\hline

retraining &
0.420 & 0.384 & -8.6 &
0.429 & 0.388 & -9.6 &
0.407 & 0.390 & -4.2 &
0.392 & 0.391 & -0.3 \\

finetune &
0.354 & 0.324 & -8.5 &
0.378 & 0.343 & -9.3 &
0.366 & 0.354 & -3.3 &
0.285 & 0.284 & -0.4 \\

EWC &
0.399 & 0.358 & -10.3 &
0.381 & 0.316 & -17.1 &
0.405 & 0.384 & -5.2 &
0.372 & 0.372 & 0.0 \\

EMR &
0.356 & 0.325 & -8.7 &
0.382 & 0.348 & -8.9 &
0.360 & 0.347 & -3.6 &
0.283 & 0.283 & 0.0 \\

LKGE &
0.418 & 0.360 & -13.9 &
0.405 & 0.341 & -15.8 &
0.407 & 0.386 & -5.2 &
0.385 & 0.385 & 0.0 \\

FMR &
0.389 & 0.339 & -12.9 &
0.358 & 0.306 & -14.5 &
0.398 & 0.371 & -6.8 &
0.382 & 0.380 & -0.5 \\

ETT-CKGE &
0.425 & 0.361 & -15.1 &
0.416 & 0.351 & -15.6 &
0.385 & 0.363 & -5.7 &
0.346 & 0.345 & -0.3 \\

DebiasedKGE &
0.413 & 0.346 & -16.2 &
0.386 & 0.310 & -19.7 &
0.388 & 0.351 & -9.5 &
0.366 & 0.365 & -0.3 \\

FastKGE &
0.392 & 0.311 & -20.7 &
0.353 & 0.286 & -19.0 &
0.397 & 0.342 & -13.9 &
0.356 & 0.353 & -0.8 \\

SAGE &
0.420 & 0.348 & -17.1 &
0.414 & 0.350 & -15.5 &
0.396 & 0.370 & -6.6 &
0.384 & 0.384 & 0.0 \\

incDE &
0.433 & 0.351 & -18.9 &
0.411 & 0.337 & -18.0 &
0.421 & 0.389 & -7.6 &
0.392 & 0.391 & -0.3 \\

\hline
\end{tabularx}
\end{table*}

\end{document}